\documentclass[letterpaper, 10 pt, conference]{ieeeconf}  

\IEEEoverridecommandlockouts                              

\usepackage[colorlinks=true,linkcolor=blue,citecolor=blue]{hyperref}%
\usepackage{lipsum} 
\usepackage{amssymb}  
\usepackage{subcaption}
\usepackage{amsmath}
\usepackage{graphicx}
\usepackage{lipsum}

\usepackage{tcolorbox}  
\tcbuselibrary{listingsutf8}  

\usepackage[ruled,vlined]{algorithm2e}

\SetKwInput{KwInput}{Input}                
\SetKwInput{KwOutput}{Output}              
\SetKwData{KwData}{Translation Estimation}

 \usepackage{nomencl}
 \usepackage{xcolor}
 \usepackage{hyperref}
 \usepackage{nomencl}
 \usepackage{hyperref}
\hypersetup{colorlinks=true,
	linkcolor=black, 
	urlcolor=blue} 
 
     \hypersetup{
  colorlinks=true,
  citecolor=blue,
  linkcolor=black,
  urlcolor=black}

\usepackage{siunitx}
\usepackage{nicematrix}
\usepackage{listings}
\usepackage{caption}
\usepackage[utf8]{inputenc}
\usepackage[T1]{fontenc}
\usepackage[english]{babel}
\usepackage{amsmath}
\usepackage{comment}
\usepackage{amsfonts}
\usepackage{subcaption}
\usepackage{multirow}
\title{\LARGE \bf
A Generalized Modeling Approach to Liquid-driven Ballooning Membranes}

\author{Mirroyal Ismayilov$^1$, Jeref Merlin$^2$, Christos Bergeles$^1$, Lukas Lindenroth$^1$
\thanks{This work was funded by the EPSRC Research Council, part of the EPSRC DTP, Grant Ref: EP/W524475/1, and supported in whole, or in part, by the EPSRC Grant Refs: EP/Z003172/1, EP/Y024281/1, the Wellcome/EPSRC Centre for Interventional and Surgical Sciences (WEISS) [203145/Z/16/Z], the Department of Science, Innovation and Technology (DSIT) and the Royal Academy of Engineering under the Chair in Emerging Technologies programme and the Enterprise Fellowship scheme. For the purpose of open access, the author has applied a CC BY public copyright licence to any author accepted manuscript version arising from this submission. The authors are with} 
\thanks{$^1$Research Department of Surgical and Interventional Engineering, School of Biomedical Engineering $\&$ Imaging Sciences, King's College London}
\thanks{$^2$UCL Hawkes Institute, Department of Medical Physics and Biomedical Engineering, University College London.}}

\begin{document}

\maketitle

\thispagestyle{empty}
\pagestyle{empty}

\begin{abstract}
Soft robotics is advancing the use of flexible materials for adaptable robotic systems. Membrane-actuated soft robots address the limitations of traditional soft robots by using pressurized, extensible membranes to achieve stable, large deformations, yet control and state estimation remain challenging due to their complex deformation dynamics.  This paper presents a novel modeling approach for liquid-driven ballooning membranes, employing an ellipsoid approximation to model shape and stretch under planar deformation. Relying solely on intrinsic feedback from pressure data and controlled liquid volume, this approach enables accurate membrane state estimation. We demonstrate the effectiveness of the proposed model for ballooning membrane-based actuators by experimental validation, obtaining the indentation depth error of
$RMSE_{h_2}=0.80\;$mm, which is $23\%$ of the indentation range and $6.67\%$ of the unindented actuator height range. For the force estimation, the error range is obtained to be $RMSE_{F}=0.15\;$N which is $10\%$ of the measured force range.
\end{abstract}

\section{INTRODUCTION}
Soft robotics is an emerging and rapidly advancing field within robotics, characterized by the development of robots made from flexible, compliant materials~\cite{Calanca2016ARobots, Tse2018SoftApplications, Damian2018InImplants}. These robots, typically composed of polymers, have a wide range of applications due to their adaptable nature. Their inherent flexibility makes them ideal for environments and tasks where conventional rigid robots might fall short, particularly in delicate or uncertain environments. However, this flexibility also introduces significant challenges in terms of control and modeling, which are not present in traditional, rigid robots~\cite{Ambaye2024SoftReview}.

While the majority of soft robots are continuum-based, which allows them to bend and flex continuously along their principal axes, controlling them is complex~\cite{Dupont2022ContinuumInterventions}. Soft continuum robots, in particular, have become popular in the healthcare sector. Their ability to navigate delicate anatomical structures has made them useful for minimally invasive procedures. These robots have demonstrated their potential to significantly enhance the safety, accuracy, and effectiveness of various medical diagnostic and therapeutic applications~\cite{Pham2020SoftFields}.
\begin{figure}[h!]
  \centering
  \includegraphics[width=1\linewidth]{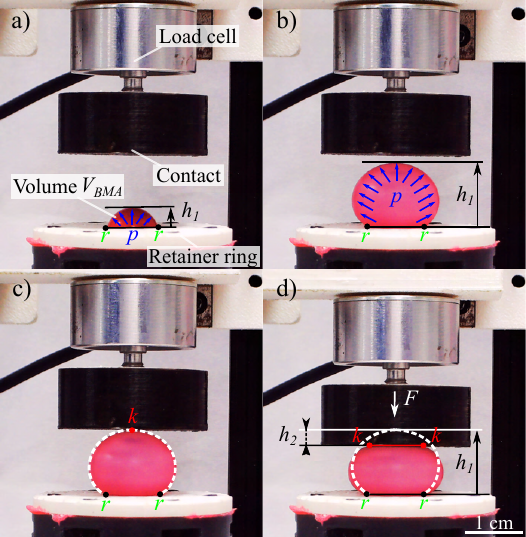}
  \caption{Overview of Ballooning Membrane Actuator (BMA) with (a) pre-ballooned membrane, (b) ballooned membrane, (c) membrane with point contact and (d) deformed membrane.}
\label{fig:four_images}
\end{figure}
Despite these advantages, soft continuum robots face limitations, such as susceptibility to buckling and low load-bearing capabilities~\cite{Herzig2021ModelMembrane}. To overcome these issues, recent research has focused on membrane-actuated soft robots. These robots leverage the superior extensibility of membranes that can expand when pressurized, much like a balloon, allowing for greater deformation without the risk of buckling. By exploiting soft materials' natural compliance and deformability, this technique creates motion through controlled expansion, opening up new possibilities for soft robotic actuation~\cite{Lindenroth2022IntrinsicInterventions,Gariya2023SoftDeformation,Lindenroth2023TowardRobots, Lindenroth2021AInjections}.

Early work on membrane expansion was pioneered by Treloar, who explored the strain distribution and shape changes of circular membranes during inflation~\cite{Treloar1944StrainsBursting, Treloar1944Stress-StrainDeformation}. Since then, extensive research has been conducted on the inflation of hyperelastic membranes, ranging from studies on simple rubber sheets to more complex models involving elastomeric membranes interacting with deformable or irregular environments~\cite{Feng1975OnMembrane, Kumar2013OnMembrane, Bouremel2018CompressionPockets}. While these models have advanced significantly, they often remain too complex to effectively integrate into control systems or observers for practical soft robotic applications.

While recent studies have advanced membrane modeling~\cite{Herzig2021ModelMembrane} and control~\cite{Shi2023ModellingLoad}, a complete state estimation of membrane actuators based solely on intrinsic feedback remains unexplored. This paper introduces a novel and generalizable method for estimating the shape and stretch of liquid-driven ballooning membranes using an ellipsoid approximation method. The ellipsoidal geometry is suited to model a wide range of ballooning profiles, which allows the axisymmetric membrane shape to adjust to deformations induced by external planar contact. Leveraging the incompressibility of the driving fluid allows for fully defining the state of the membrane under external load through the induced liquid volume and pressure feedback alone, a capability that has not been previously addressed in the literature. In the following sections, we derive the modeling framework and validate it experimentally using a hyperelastic silicone rubber membrane subjected to time-varying external contacts.

    

\section{METHODS}
\subsection{Related Work}
Recent research in membrane-based actuation reveals similar approaches for estimating membrane stretch across studies. Most commonly, this involves calculating principal stretches by integrating over infinitesimally small membrane segments, offering a detailed yet computationally intensive method~\cite{Zhou2018AnMembrane, Wang2017AnomalousActuation, Shi2022ModellingFingertip, Shi2023ModellingLoad}. However, this technique is primarily applicable to small expansion profiles, as it tends to produce errors in scenarios involving high strain. On the other hand, Herzig et al. \cite{Herzig2021ModelMembrane} proposed a geometric model for highly extensible membranes where the deformed membrane is approximated as a spherical segment under specific conditions.  The approach takes many simplifications describing the membrane's deformation configurations regardless of whether the external force is applied or not. While various methods are employed to estimate stretch and strain energy, the core concept involves balancing the total potential energy of the inflated membrane based on the principle of minimum potential energy. When a compressible fluid is used as the driving medium, additional formulations are required to account for compressibility, making it more complex to relate input volume with actuator shape, thus complicating state estimation~\cite{Herzig2021ModelMembrane,Shi2023ModellingLoad}. In contrast, using liquid as a driving medium eliminates the need for these equations, simplifying the modeling process.

As shown in Fig.~\ref{fig:four_images}, the state variables are selected to be $(h_1,\;h_2,\; F)$, where $h_1$ represents the virtual unindented height of the membrane from the baseline assuming that the membrane is in the non-contact phase and $h_2$ is the actual depth to which the membrane is pushed down from the unindented height $h_1$ due to the applied external force $F$ when the membrane is in the contact phase.

\begin{figure}[thpb]
  \centering
  \includegraphics[width=1\linewidth]{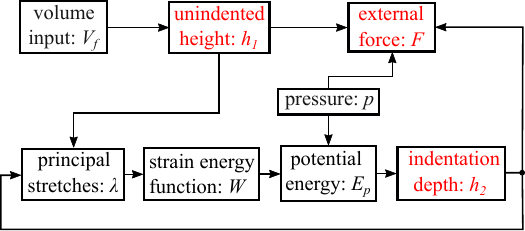}
  \caption{Flow diagram: intrinsic quasi-static modeling, the state variables are indicated in red color.}
  \label{fig: flow_chart}
\end{figure}

As illustrated in Fig.~\ref{fig: flow_chart}, the generic way to compute state variables of liquid-driven ballooning membrane actuators (BMAs) is as follows:
The model starts with a volume input representing the fluid inside the membrane. From the volume, the unindented height $h_1$ is calculated. At this stage, the membrane takes on a specific shape, which needs to be calculated based on geometric assumptions. With the geometry determined, the membrane’s deformation can be calculated. The principal stretches $\lambda$ refers to the change in lengths along the membrane's principal directions, which are required for further energy calculations. The strain energy function $W$ is computed using the material's properties based on the principal stretches. The total potential energy 
$E_p$ of the BMA represents the work done to deform the membrane elastically and combines three components: the strain energy stored in the membrane due to deformation, the work done by the relative pressure $p$ to the environment over the volume of fluid, and the work from an external force 
$F$ acting over the indentation height difference $h_1-h_2$.
\begin{equation}
    E_p=\int_{V_m} W \,dV-\int_{V_f} p \,dV + F(h_1-h_2)
    \label{Eq:E_p_prior_art}
\end{equation}
where $V_m$ is the membrane's volume and $V_f$ is the fluid volume within the membrane.

\subsection{Proposed Modeling Approach}
In contrast to the approach in \cite{Herzig2021ModelMembrane}, where air was utilized as the driving medium for generating deformation, our study focuses on liquid as the primary medium for actuation. 
Liquid as the driving medium results in different deformed membrane configurations due to its incompressibility.

In our approach, we observed that the inflated membrane shape typically deviates from a perfect sphere, resembling an ellipsoid instead. To account for this, we approximate the membrane as a perfect ellipsoid to compute stretch during actuation. While the actual deformation likely deviates from this ideal, this approximation enhances the generalizability of our approach. 

\begin{figure}[t!] 
  \begin{subfigure}[b]{0.5\linewidth}
    \centering
    \includegraphics[width=0.9\linewidth]{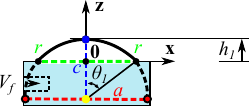} 
    \caption{} 
    \label{Fig:Model_0} 
    \vspace{1ex}
  \end{subfigure}
  \begin{subfigure}[b]{0.5\linewidth}
    \centering
    \includegraphics[width=0.9\linewidth]{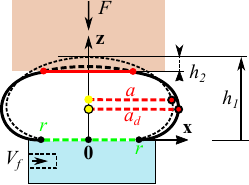} 
    \caption{} 
    \label{Fig:Model_1} 
    \vspace{1ex}
  \end{subfigure} 
  \vspace{1px}
  
  \begin{subfigure}[b]{0.5\linewidth}
    \centering
    \includegraphics[width=0.9\linewidth]{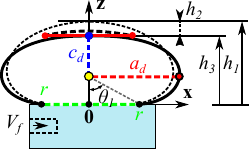} 
    \caption{} 
    \label{Fig:Model_2} 
  \end{subfigure}
  \begin{subfigure}[b]{0.5\linewidth}
    \centering
    \includegraphics[width=0.9\linewidth]{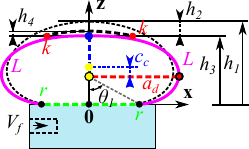} 
    \caption{} 
    \label{Fig:Model_3} 
  \end{subfigure} 
  \vspace{1px}
  
  \caption{Modeling of BMA: (a) Pre-ballooned configuration of the membrane (non-contact phase). (b) Interaction forces during external contact. (c) Shape estimation of BMA under contact. (d) Regions of ellipsoid components define the estimated stretch of the membrane.}
  \label{fig: Modeling} 
\end{figure}
The actuator volume $V_{BMA}$ comprises injected liquid volume $V_f$ and membrane volume $V_m$:
\begin{equation}
    V_{BMA}= V_f+V_m
    \label{eq:V_a_st}
\end{equation}
The membrane volume $V_m$ is computed as below:
\begin{equation}
    V_m=r^2\pi\; t_i
    \label{eq:V_m_st}
\end{equation}
where $t_i$ is the initial thickness of the membrane that is assumed to be uniform, and $r$ is the inner radius of the retaining ring.

Geometrically, the volume of the actuator can also be defined as:
\begin{equation}
    V_{BMA}= \frac{4}{3} \pi  a^2 c-V_{b}
    \label{eq:V_f_st}
\end{equation}
where $a$ and $c$ are major and minor axes of the ellipsoid and $V_b$ is the virtual cap volume of the ballooned membrane below the retainer ring, which completes the ellipsoid.
The virtual cap volume can be computed as below:
\begin{equation}
  V_{b}= \frac{{{a}^{2}} \left( 3 c-h_b\right)  {{h_b}^{2}} \pi}{3 {{c}^{2}}}
  \label{Eq:V_b}
\end{equation}
where $h_b$ corresponds to the virtual cap height below the retainer ring clamping the membrane:
\begin{equation}
    h_b=2c-h_1
    \label{Eq:h_b}
\end{equation}
Using (\ref{Eq:h_b}) and (\ref{Eq:V_b}), (\ref{eq:V_f_st}) can be rewritten to:
\begin{equation}
    V_{BMA}=-\frac{{{a}^{2}} {{\left( 2 c-{h_1}\right) }^{2}} \left( {h_1}+c\right)  \pi}{3 {{c}^{2}}}+\frac{4 \pi {{a}^{2}} c}{3}
    \label{Eq:V_f_scnd}
\end{equation}
Rearranging (\ref{Eq:V_f_scnd}), the major axis $a$ can be expressed as:
\begin{equation}
    a=\frac{\sqrt{3} c \sqrt{\frac{V_{BMA}}{\pi \left( 3 c-{h_1}\right)  }}}{{h_1}}
    \label{Eq:a_first}
\end{equation}
Another equality comes from the boundary condition regarding the retainer ring:
\begin{equation}
    A_i={{a}^{2}} \left( 1-{{\left( 1-\frac{h_b}{c}\right) }^{2}}\right)  \pi
    \label{Eq:A_i_1}
\end{equation}
where $A_i$ is the initial surface area of the membrane and is defined as:
\begin{equation}
    A_i=\pi r^2
    \label{Eq:A_i_2}
\end{equation}

By solving the system of equations (\ref{Eq:A_i_1}) and (\ref{Eq:A_i_2}), and substituting the major axis with (\ref{Eq:a_first}) and the virtual cap height with (\ref{Eq:h_b}), the minor axis is expressed as:
\begin{equation}
c=\frac{{{{h_1}}^{2}} \pi {{{r}}^{2}}-3 V_{BMA} {h_1}}{3 {h_1} \pi {{{r}}^{2}}-6 V_{BMA}}
\label{Eq:c}
\end{equation}
The minor axis equation given in (\ref{Eq:c}) is substituted in (\ref{Eq:a_first}) to compute the major axis of the ellipsoid ballooned membrane:
\begin{equation}
    a=\frac{
    \sqrt{-\frac{{h_1} \pi {{{r}}^{2}}-2 V_{BMA}}{{h_1} \pi}} \left( \sqrt{3} {h_1} \pi {{{r}}^{2}}-{{3}^{\frac{3}{2}}} V_{BMA}\right) 
    }
    {3 {h_1} \pi {{{r}}^{2}}-6 V_{BMA}}
    \label{Eq:a}
\end{equation}
Knowing the injected volume and the unindented membrane height, the ellipsoid can be fully reconstructed. Since the injected volume is already known, unindented height is the only unknown parameter that needs to be linked to the injected volume. An extra equation is needed to build a functional relationship between the known injected volume and the corresponding unindented height of the membrane.
Unindented height estimation from the injected volume can be described as:
\begin{equation}
 h_1=f(V_f)
 \label{Eq:h_1}
\end{equation}
where $f$ is a higher-order polynomial.
Once the fluid volume to unintended height is established, the unindented ellipsoid membrane shape can be fully estimated.

Once an external force is applied to the membrane, the ellipsoid geometry is deformed iteratively.
The main geometric parameters for deformed ellipsoid approximation are the deformed major and minor axes, denoted as $a_d$ and $c_d$, respectively. The key assumption is reconstructing the new ellipsoid geometry with iterative increments based on the adjusted deformed height and the actuator volume.
\begin{equation}
c_d=\frac{{{{h_3}}^{2}} \pi {{{r}}^{2}}-3 V_{BMA} {h_3}}{3 {h_3} \pi {{{r}}^{2}}-6 V_{BMA}}
\label{Eq:c_d}
\end{equation}
where $h_3$ corresponds to the membrane's deformed height during contact:
\begin{equation}
    h_3=h_1-h_{2,\;t-1}
    \label{Eq:h_3}
\end{equation}
in the equation above, $h_{2,\;t-1}$ is the total indentation of the membrane computed in the previous iteration.  \\
The major axis of the deformed membrane is calculated as follows:
\begin{equation}
    a_d=\frac{\sqrt{-\frac{{h_3} \pi {{{r}}^{2}}-2 V_{BMA}}{{h_3} \pi}} \left( \sqrt{3} {h_3} \pi {{{r}}^{2}}-{{3}^{\frac{3}{2}}} V_{BMA}\right) }{3 {h_3} \pi {{{r}}^{2}}-6 V_{BMA}}
    \label{Eq:a_d}
\end{equation}

As illustrated in Fig.~\ref{fig: Modeling}, external contact causes the membrane to be both sliced and displaced along the direction of the applied force. This results in an adjusted contact area as the membrane shifts away from the location where the force is applied. The displacement, representing a shift in the central axis of the membrane’s geometry due to deformation, can be determined as follows:
\begin{equation}
c_c=c-c_d
\label{Eq:c_c}
\end{equation}
Deriving the principal stretch of the ellipsoid membrane involves several key steps, starting with identifying membrane regions based on the injected volume and corresponding indentation depth.
\begin{figure}[t!]
  \centering
  \vspace{-25px}
  \includegraphics[width=1\linewidth]{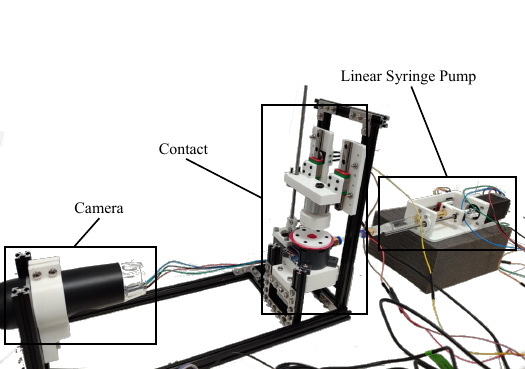}
  \caption{Experimental setup for model validation}
  \label{Fig:Experimental_setup}
\end{figure}
The ellipse's perimeter is computed utilizing integral boundaries as a function of the specific dimensions and characteristics of the ellipsoidal geometry.
\begin{equation}
{{\theta }_1}=\arctan\left( \frac{{r}}{\left| {h_3}-c_d\right| }\right)
\label{Eq:theta_1}
\end{equation}

\begin{eqnarray}
L = \left\{ 
\begin{array}{ll}
    \int_{0}^{\pi-\theta_1} M \,dt , & \quad \mathrm{if}\; h_3>c_d \\
   \int_{0}^{\theta_1} M \,dt, & \quad \mathrm{ otherwise},
\end{array}
\right.
\label{Eq:L}
\end{eqnarray}
\begin{equation}
M=\sqrt{a_d^2\sin^2(t)+c_d^2\cos^2(t)}
\end{equation}
The complete stretch profile of the membrane is obtained by:
\begin{equation}
    \lambda=\frac{L}{r}
    \label{Eq:lambda}
\end{equation}
We can define the Cauchy-Green invariant $I_1$ in this particular case as a function of the stretch $\lambda$:
\begin{equation}
    I_1=\lambda^2+\frac{2}{\lambda}
    \label{Eq:I_1}
\end{equation}
Once the stretch estimation is established due to highly non-linear observed behavior, a higher-order function is selected; thus, the Yeoh, 6th order model is selected:
\begin{equation}
W=\sum_{n=0}^{6} 2\; (\lambda-\lambda^{-2})\cdot n\cdot C_n\cdot (I_1-3)^{n-1}
\label{Eq:W}
\end{equation}
where $C_n$ constant variables are the material parameters of the membrane for a given $n^{th}$ order model~\cite{Marechal2021TowardRobotics}.

Under the minimum potential energy assumption, the total potential energy equality is used:
\begin{equation}
    E_p=\int_{V_{fm}} W \,dV-\int_{V_f} p \,dV + F\; h_3
    \label{Eq:E_p}
\end{equation}

The membrane material is considered incompressible. However, during deformation, only the free inflation region of the membrane $V_{fm}$, which represents the portion not in contact with the external contact piece, is included in the energy balance equation. This is because the membrane in contact with the external object is assumed not to store strain energy. The volume of the membrane corresponding to the free inflation region is defined as:
\begin{equation}
    V_{fm}=V_m-k^2\pi\; t_m
    \label{Eq:V_fm}
\end{equation}
where $t_m$ is the thickness of the inflated membrane:
\begin{equation}
    t_m=\frac{t_i\; r^2}{L^2}
    \label{Eq:t_m}
\end{equation}
The contact point $k$ is computed by slicing the unindented membrane based on the total indentation depth computed in the previous iteration.
\begin{equation}
k= \frac{a \sqrt{2 c\;(h_{2,\;t-1}-c_c)-(h_{2,\;t-1}-c_c)^2}}{c}
\label{Eq:k}
\end{equation}
By rearranging the Eq.~\ref{Eq:E_p}, the external planar force is estimated:
\begin{equation}
    F=\frac{V_f\;p-V_{fm}W}{h_3}
    \label{Eq:F}
\end{equation}
The slice-induced indentation depth of the deformed membrane can be computed as follows:
\begin{equation}
h_4=\frac{-(c \sqrt{\pi^2 a^2\;p^2 -\pi F p }-\pi a\;c\;p )}{\pi a\;p }
\label{Eq:h_4}
\end{equation}
Finally, the current total indentation depth of the membrane is updated by combining slice-induced and motion-induced indentations:
\begin{equation}
h_{2,\;t}=h_4+c_c
\label{Eq:h_2}
\end{equation}
To ensure stable state estimation, the indentation depth is filtered to remain within the unindented height limit.
\begin{algorithm}[thpb]
\caption{Proposed Modeling Algorithm}
\DontPrintSemicolon
  \nl Set material parameters: $C$\;
  \nl Set height fitting function: $f$\;
  \KwInput\;
  Injected volume: $V_{f}$\;
  Sensed internal pressure: $p$\;
  \KwOutput\;
  Estimated unintended ellipsoid height: $h_{1}$\;
  Estimated indentation depth: $h_{2}$\;
  Estimated deformed ellipsoid height: $h_{3}$\;
  Estimated external force: $F$\;
   \While{$V_f$}
   {
     \begin{tcolorbox}[colback=white!10, colframe=black!40, boxrule=0.5pt, width=0.8\columnwidth, left=1pt, right=1pt, top=1pt, bottom=1pt] 
   \textbf{Data Acquisition}\;
   \nl Obtain sensor pressure value: $p$\;
   \nl Obtain injected volume: $V_{f}$\;
    \end{tcolorbox}

         \begin{tcolorbox}[colback=brown!10, colframe=brown!40, boxrule=0.5pt, width=0.8\columnwidth, left=1pt, right=1pt, top=1pt, bottom=1pt] 
   \textbf{Height Fitting}\;
   \nl Compute unindented height $h_{1}$ using (\ref{Eq:h_1})\;
    \end{tcolorbox}
   
  \begin{tcolorbox}[colback=red!10, colframe=red!40, boxrule=0.5pt, width=0.8\columnwidth, left=1pt, right=1pt, top=1pt, bottom=1pt] 
   \textbf{Shape Estimation}\;
   \nl Reconstruct the unindented membrane shape using (\ref{Eq:c}) and (\ref{Eq:a})\;
   \nl Compute deformed height $h_3$ using (\ref{Eq:h_3})\;
   \nl Reconstruct the deformed membrane shape using (\ref{Eq:c_d}) and (\ref{Eq:a_d})\;
   \nl Compute push-induced indentation $c_c$ using (\ref{Eq:c_c})\;
   \nl Compute contact point $k$ using (\ref{Eq:k})\;
   \end{tcolorbox}

  \begin{tcolorbox}[colback=green!10, colframe=green!40, boxrule=0.5pt, width=0.8\columnwidth, left=1pt, right=1pt, top=1pt, bottom=1pt] 
   \textbf{Pressure Estimation}\;
   \nl Compute the stretch $\lambda$ using (\ref{Eq:lambda})\;
   \nl Compute strain energy density $W$ using (\ref{Eq:W})\;
   \end{tcolorbox}
   
   \begin{tcolorbox}[colback=cyan!10, colframe=cyan!40, boxrule=0.5pt, width=0.8\columnwidth, left=1pt, right=1pt, top=1pt, bottom=1pt] 
   \textbf{State Estimation}\;
   \nl Compute external force $F$ using (\ref{Eq:F})\;
   \nl Compute slice-induced indentation depth $h_4$ using ($\ref{Eq:h_4}$)\;
   \nl Update total indentation  depth $h_{2,\;t}$ using ($\ref{Eq:h_2}$)\;
   \end{tcolorbox}
   }
\end{algorithm}

\begin{figure}[thpb] 
  \begin{subfigure}[b]{0.5\linewidth}
    \centering
    \includegraphics[width=1\linewidth]{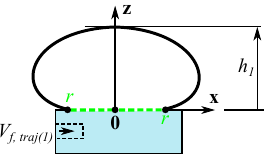} 
    \caption{} 
    \label{Fig:traj_0} 
    \vspace{1ex}
  \end{subfigure}
  \begin{subfigure}[b]{0.5\linewidth}
    \centering
    \includegraphics[width=1\linewidth]{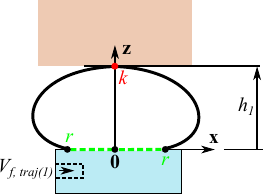} 
    \caption{} 
    \label{Fig:traj_1} 
    \vspace{1ex}
  \end{subfigure} 
  \vspace{1px}
  
  \begin{subfigure}[b]{0.5\linewidth}
    \centering
    \includegraphics[width=1\linewidth]{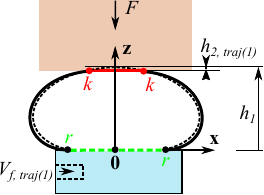} 
    \caption{} 
    \label{Fig:traj_2} 
  \end{subfigure}
  \begin{subfigure}[b]{0.5\linewidth}
    \centering
    \includegraphics[width=1\linewidth]{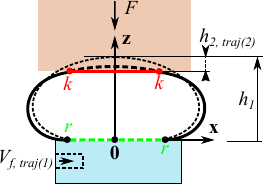} 
    \caption{} 
    \label{Fig:traj_3} 
  \end{subfigure} 
  \vspace{1px}
  
  \caption{Trajectory sequencing on the ballooning membrane: (a) The formed geometry with selected volume trajectory (non-contact phase), (b) Interaction of the formed membrane with point contact assuming no force or indentation applied, and (c) The first indentation trajectory point for a given volume trajectory. (d) The second indentation trajectory point for the same volume.}
  \label{fig: Traj Sequencing} 
\end{figure}
\subsection{Experimental Setup}
To validate the proposed model, an experimental setup was developed as shown in Fig.~\ref{Fig:Experimental_setup}, incorporating a microcontroller (ESP32-C3, Espressif, Shanghai, China) for coordination and a laptop to handle tasks via a ROS2 interface. A pump precisely injects water into an Ecoflex 00-50 (Smooth-On, Inc.) membrane actuator (0.5 mm thickness, 5 mm ring radius) using a 2.25 ml syringe (Borosilicate Glass Syringe, Bitomic, China), driven by a stepper motor (Nema 11, OSM TECHNOLOGY CO., LTD., China) regulated with a motor controller (TMC5160-BOB, TRINAMIC Motion Control GmbH $\&$ Co. KG., Germany). A pressure sensor (MPRLS 0-25 PSI, Adafruit, US) monitors chamber pressure, while a camera (RS PRO USB Digital Microscope 5M pixels, RS Components, China) captures real-time deformation images for model validation. Additionally, a force sensor (DYMH-103 0-50 kg, Shenzhen, China), integrated with the stepper motor, measures the force applied as the motor targets indentation on the membrane through a contact piece.

\section{Results}
To compute the accuracy of estimated parameters, RMSE (Root Mean Square Error) is computed:
\begin{equation}
RMSE=\sqrt{\frac{\sum_{n=1}^{N}(\hat{y}-y)^2}{N}}
\end{equation}
where $y$ is the measured variable, $\hat{y}$ is the estimated variable, and
$N$ is the number of compared data points.
\subsection{Height fitting and Shape Estimation} 
\begin{figure}[t!]
      \centering
      \includegraphics[width=1\linewidth]{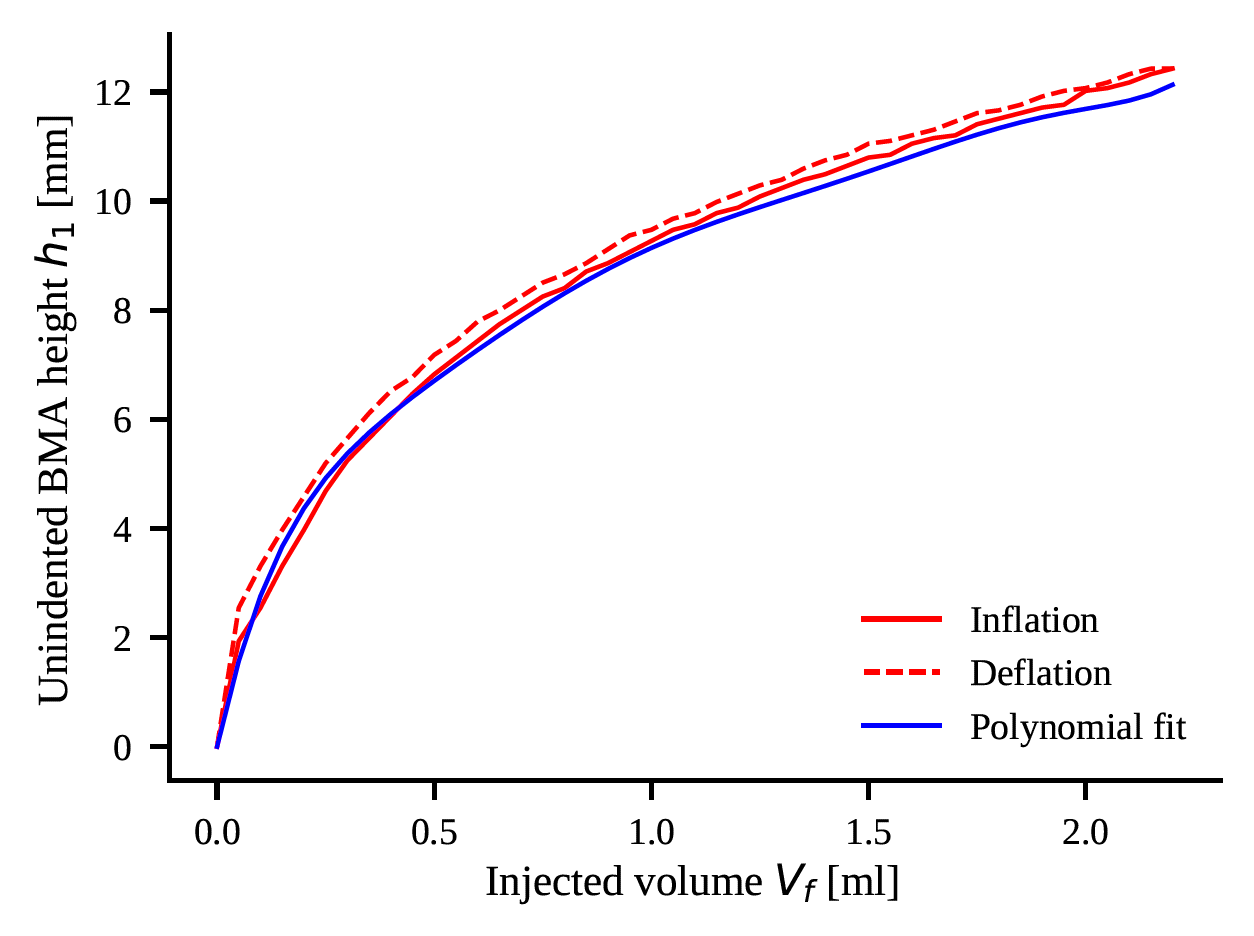}
      \caption{Polynomial ($7^{th}$ order) height fitting based on camera feedback}
      \label{Fig: Height fitting}
\end{figure}

\begin{figure}[t!] 
  \begin{subfigure}[b]{0.5\linewidth}
    \centering
    \includegraphics[width=0.9\linewidth]{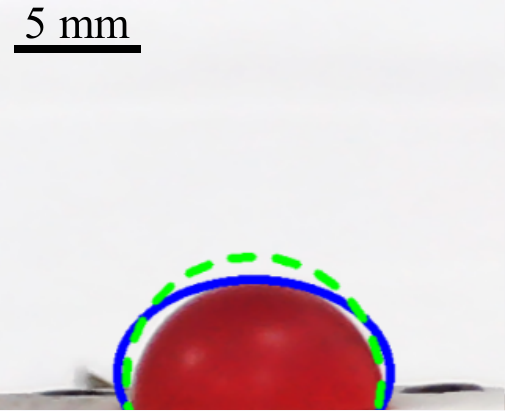} 
    \caption{} 
    \label{Fig:capture 1} 
    \vspace{0.5ex}
  \end{subfigure}
  \begin{subfigure}[b]{0.5\linewidth}
    \centering
    \includegraphics[width=0.9\linewidth]{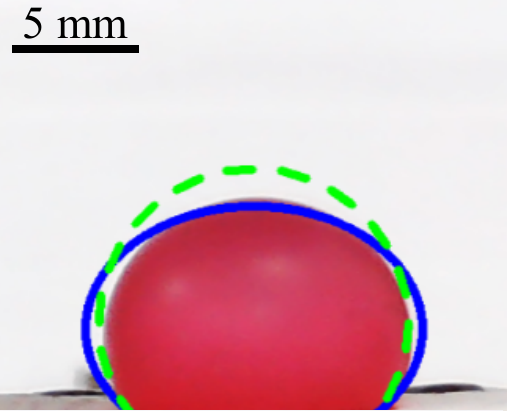} 
    \caption{} 
    \label{Fig:capture2} 
    \vspace{0.5ex}
  \end{subfigure} 
  \vspace{1px}
  
  \begin{subfigure}[b]{0.5\linewidth}
    \centering
    \includegraphics[width=0.9\linewidth]{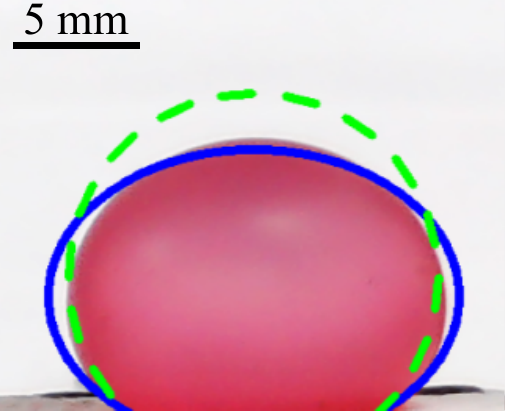} 
    \caption{} 
    \label{Fig:capture3} 
  \end{subfigure}
  \begin{subfigure}[b]{0.5\linewidth}
    \centering
    \includegraphics[width=0.9\linewidth]{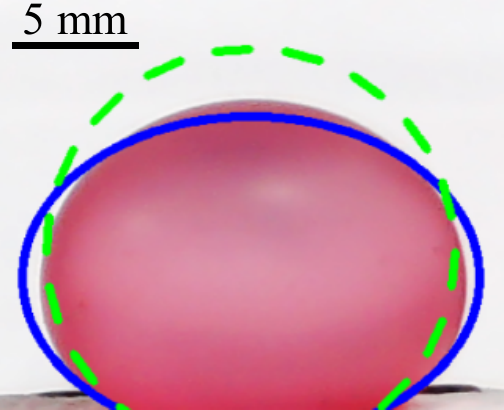} 
    \caption{} 
    \label{Fig:capture4} 
  \end{subfigure} 
  \vspace{1px}
  
  \caption{Estimated ellipsoid (blue) overlaid on membrane images for pre-ballooned and ballooned states. For comparison, a sphere is fitted under consideration of the same fluid volume (green).}
  \label{Fig: Experimental results: shape}
\end{figure}

The mapping of fluid volume to unindented height is established via camera feedback. As observed in Fig.~\ref{Fig: Height fitting}, stress relaxation of the selected material due to deformation contributes to the hysteresis in the membrane's inflation and deflation height profiles. Height fitting is then performed based on the mean values of these measured heights. To predict the unintended virtual height, a 7th-order polynomial function is fitted and used.
To compare the actual shape of the membrane with the proposed ellipsoid approximation, the reconstructed shapes are plotted against the actual membrane geometry, as illustrated in ~Fig.~\ref{Fig: Experimental results: shape}. For reconstructing the modeling geometries, input volume $V_f$ is used, and then geometries are determined and displayed in image format.

\subsection{Model Validation}
\textit{Pressure Estimation}: Initially the membrane is actuated across multiple trajectory points $V_{f,\;traj}$ while estimated pressure from modeling is compared with sensor pressure values, all conducted without any external contact applied to the membrane. 

\begin{figure}[t!]
      \centering
      \includegraphics[width=1\linewidth]{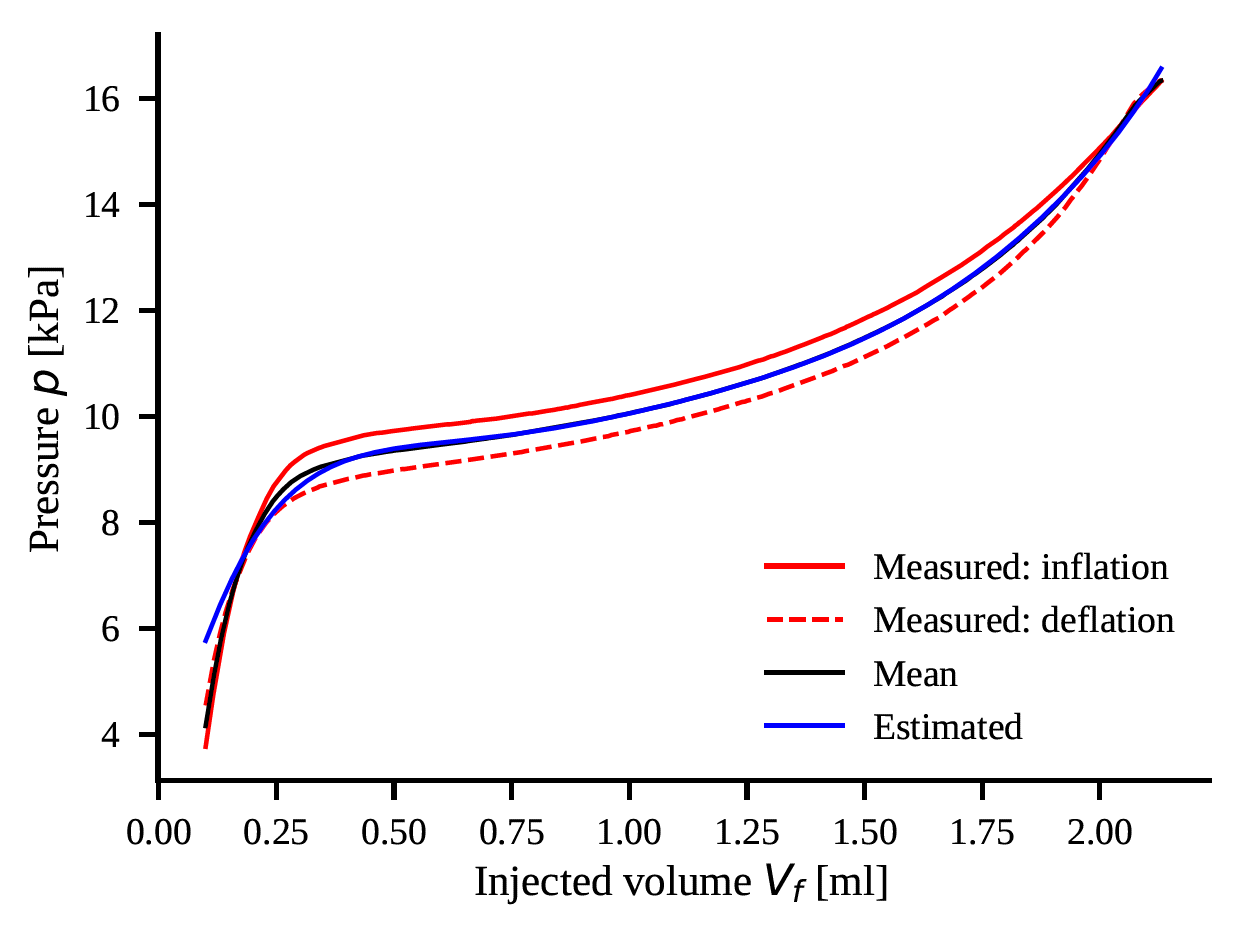}
      \caption{Pressure Estimation}
      \label{Fig: Pressure Estimation}
\end{figure}

The pressure estimation is depicted in Fig.~\ref{Fig: Pressure Estimation}, where the estimated pressure deviates from the mean pressure during the early expansion stage. This problem could potentially come from the selected material model, which struggles to capture the highly non-linear membrane material behavior. 

For the pressure estimation, the error range is computed to be $RMSE_p=168.01\;$Pa which makes up $1.03\%$ of the measured pressure range.

The error measurements are carried out for the hysteresis behavior of the membrane and $RMSE_{p,\; hyster}=323.36\;$Pa is computed for both inflation and deflation curves in comparison with mean pressure which corresponds to $1.98\%$ of the pressure range for both computed values.

\textit{State Estimation}: As illustrated in Fig.~\ref{fig: Traj Sequencing}, state estimation and modeling are validated through multiple injected volume trajectory points 
$V_{f,\;traj}$ and indentation depth trajectory points $h_{2,\;traj}$. For each unindented height trajectory point, all corresponding indentation depth trajectory points must be achieved.
The main error deviation is computed within the time frame when there is an actual indentation applied to the membrane For this region, the indentation depth error range is computed to be
$RMSE_{h_2}=0.80\;$mm which is $23\%$ of the indentation range and $6.67\%$ of the unindented actuator height. For the force estimation, the error range is
$RMSE_{F}=0.15\;$N which is $10\%$ of measured force range.

\begin{figure}[thpb]
      \centering
      \includegraphics[width=\linewidth]{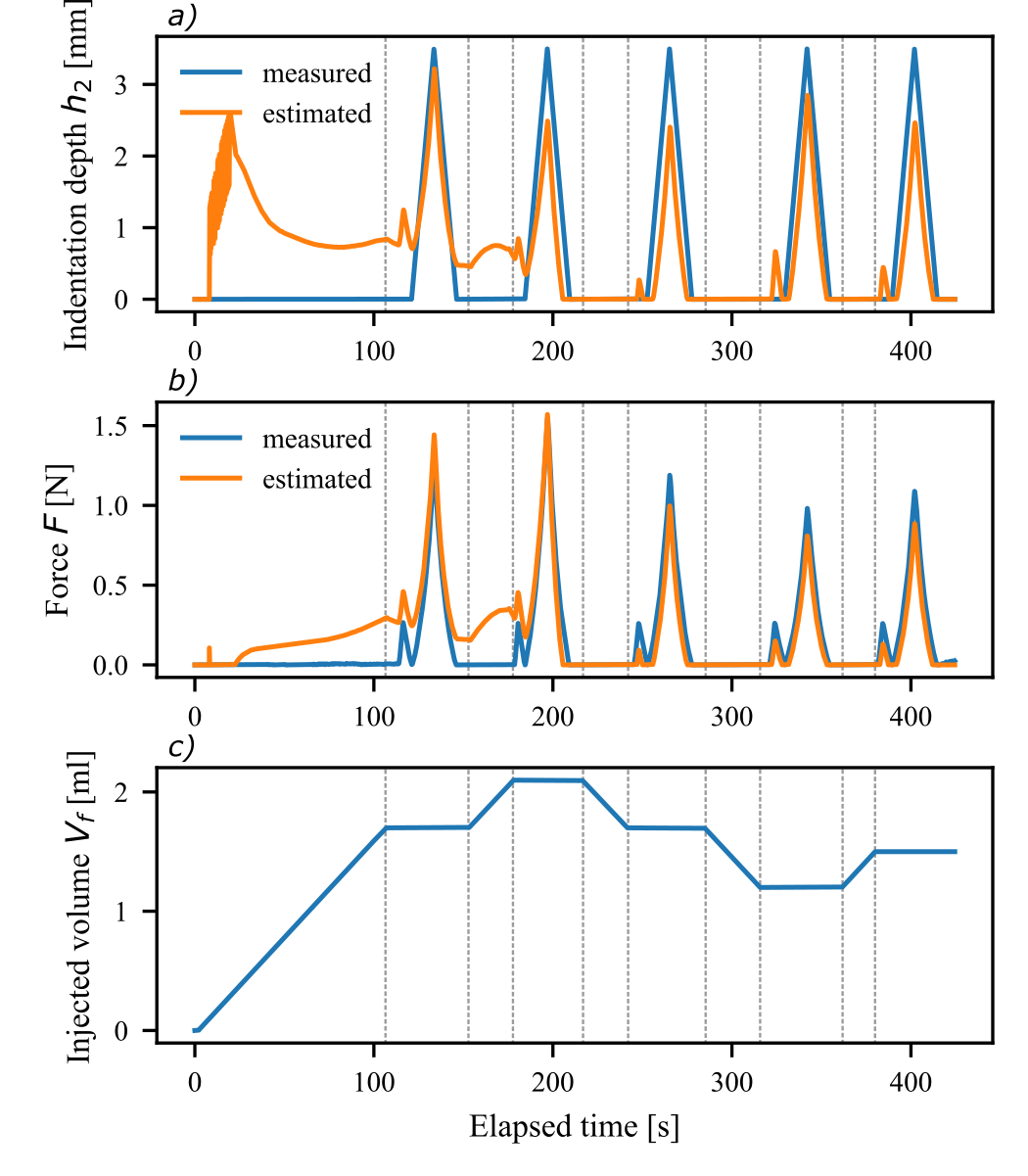}
      \caption{State Estimation,  a) Comparison of indentation depth estimation with the actual indentation depth, b) Force estimation plotted against the actual force feedback from the load cell, c) Injected liquid volume to the ballooning membrane}
      \label{Fig: State Estimation}
\end{figure}

As shown in Fig.~\ref{Fig: State Estimation}, the initial part of the state estimation is shown to be prone to more error deviation. The error measurements for state variables from the start of the experiment until the first actual indentation are computed. For the initial region, the indentation depth error is computed to be
$RMSE_{h_2}=1.11\;$mm. For the force estimation, the deviation error is calculated to be
$RMSE_{F}=0.16\;$N.

\section{DISCUSSIONS}
This paper presents and validates a novel, generalizable modeling approach for liquid-driven ballooning actuators. The proposed method relies solely on intrinsic real-time membrane feedback, including pressure and the volume of injected liquid as an input.

It is important to note that a limitation of the proposed model appears in the early stages of shape estimation, specifically within the initial 0–0.1 ml liquid range. At this stage, volume imbalance occurs because the membrane volume $V_m$ is notably larger than the injected liquid volume $V_f$, and only a portion of the membrane volume contributes to the actuator’s volume. This discrepancy, caused by height fitting, affects the accuracy of the ellipsoid shape estimation. To mitigate this issue later, modeling is omitted for the 0–0.1 ml injection range.

As illustrated in Fig.~\ref{Fig: State Estimation}, the model accurately captures trends in state estimation. However, due to the hysteresis properties of the hyperelastic membrane material, certain inaccuracies arise: the model tends to overestimate the external force during inflation and underestimate state variables during deflation. The initial spikes in force and indentation depth estimation come from force homing, which is used for determining the absolute zero indentation depth from where the indentation depth is measured.

During state estimation, it is observed that the modeling is sensitive to errors in the pre-ballooned membrane state. When the membrane is minimally inflated, even slight estimation errors result in large deviations in stretch ratio, leading to instabilities. This issue is further amplified by the selected material model, which does not perfectly capture the strain energy stored in the membrane. To avoid estimation divergence and unexpected sensor spikes, we implemented a filter to limit sensory feedback.

The errors in state estimation are relatively low for force and indentation depth estimation. The modeling could potentially be used for developing controllers and observers for ballooned-membrane-actuated soft robots when they are deployed for planar contact scenarios. However, to incorporate the modeling into non-planar or angled contact cases, the work needs to be extended.

One limitation of the proposed modeling approach is fitting the injected liquid volume to the unindented membrane height. While this step is essential for accurately estimating the expanded membrane shape, simplifying this process could make it avoidable in future iterations. Another limitation, as mentioned earlier, is the hysteresis behavior of the membrane material; incorporating hysteresis modeling could address this issue in future work.

In conclusion, the proposed model lays a foundation for future exploration into the development of controllers and observers for ballooned-membrane-actuated soft robots.
\bibliographystyle{ieeetr}
\bibliography{bibliography}
\end{document}